\documentclass[journal]{IEEEtran}
\usepackage{amsmath,amsfonts}
\usepackage{algorithmic}
\usepackage{algorithm}
\usepackage{array}
\usepackage[caption=false,font=normalsize,labelfont=sf,textfont=sf]{subfig}
\usepackage{textcomp}
\usepackage{stfloats}
\usepackage{url}
\usepackage{verbatim}
\usepackage{graphicx}
\usepackage{amssymb}
\usepackage{tikz}
\usepackage{amsthm}
\usepackage{enumitem}
\usepackage[numbers,sort&compress]{natbib}
\usepackage{xcolor}
\usepackage{eso-pic}

\definecolor{infernolow}{HTML}{160b39}
\definecolor{infernomid}{HTML}{bb3754}
\definecolor{infernohigh}{HTML}{f5d745}

\newcommand{\candidatemarkerlow}{\textcolor{infernolow}{$\bullet$}}
\newcommand{\candidatemarkermid}{\textcolor{infernomid}{$\bullet$}}
\newcommand{\candidatemarkerhigh}{\textcolor{infernohigh}{$\bullet$}}
\newcommand{\groundtruthmarker}{\textcolor{magenta}{$\boldsymbol{\times}$}}
\newcommand{\datamarker}{\textcolor{red}{$\circ$}}

\newtheorem{theorem}{Theorem}
\newtheorem{lemma}[theorem]{Lemma}
\newtheorem{remark}[theorem]{Remark}

\AddToShipoutPictureBG*{%
    \AtPageUpperLeft{%
        \setlength\unitlength{1in}%
        \hspace*{\dimexpr0.5\paperwidth\relax}
        \makebox(0,-0.8)[c]{%
            \begin{tabular}{c}
                J.S. van Hulst \emph{et al.}, ``Bridging the Simulation-to-Reality Gap in Electron Microscope Calibration via\\
                VAE-EM Estimation.'' This work has been submitted to the IEEE for possible publication. Copyright\\
                may be transferred without notice, after which this version may no longer be accessible.
            \end{tabular}%
        }%
    }%
}%

\AddToShipoutPictureBG*{%
\AtPageUpperLeft{%
\setlength\unitlength{1in}%
\hspace*{\dimexpr0.5\paperwidth\relax}
\makebox(0,-21.3)[c]{
\footnotesize
\begin{tabular}{c c}
\copyright~The Authors
\end{tabular}}}}

\begin{document}

\title{Bridging the Simulation-to-Reality Gap in Electron Microscope Calibration via VAE-EM Estimation}

\author{Jilles S. van Hulst$^{*}$,
        W.P.M.H. (Maurice) Heemels,
        and~Duarte J. Antunes
\thanks{The authors are with the Control Systems Technology Section, Department of Mechanical Engineering, Eindhoven University of Technology, 5600 MB Eindhoven, The Netherlands.}
\thanks{*Corresponding author. Email: {\tt j.s.v.hulst@tue.nl}.}
\thanks{This research is partly funded by the ITEA4 20216 ASIMOV project. The ASIMOV activities are supported by the Netherlands Organisation for Applied Scientific Research TNO and the Dutch Ministry of Economic Affairs and Climate (project number: AI211006). The research leading to these results is partially funded by the German Federal Ministry of Education and Research (BMBF) within the project ASIMOV-D under grant agreement No. 01IS21022G [DLR], based on a decision of the German Bundestag.}
\thanks{This research is also funded by the research project entitled \emph{Learning in Motion}, a collaboration between the Eindhoven University of Technology and several industry partners. This project is co-financed by Holland High Tech, top sector High-Tech Systems and Materials, with a PPP innovation subsidy for public-private partnerships for research and development.}}

\markboth{}%
{van Hulst \MakeLowercase{\textit{et al.}}: Bridging the Simulation-to-Reality Gap in STEM Calibration via VAE-EM}

\maketitle

\begin{abstract}
Electron microscopy has enabled many scientific breakthroughs across multiple fields. A key challenge is the tuning of microscope parameters based on images to overcome optical aberrations that deteriorate image quality.
This calibration problem is challenging due to the high-dimensional and noisy nature of the diagnostic images, and the fact that optimal parameters cannot be identified from a single image.
We tackle the calibration problem for Scanning Transmission Electron Microscopes (STEM) by employing variational autoencoders (VAEs), trained on simulated data, to learn low-dimensional representations of images, whereas most existing methods extract only scalar values. We then simultaneously estimate the model that maps calibration parameters to encoded representations and the optimal calibration parameters using an expectation maximization (EM) approach. This joint estimation explicitly addresses the simulation-to-reality gap inherent in data-driven methods that train on simulated data from a digital twin. We leverage the known symmetry property of the optical system to establish global identifiability of the joint estimation problem, ensuring that a unique optimum exists.
We demonstrate that our approach is substantially faster and more consistent than existing methods on a real STEM, achieving a $2\times$ reduction in estimation error while requiring fewer observations.
This represents a notable advance in automated STEM calibration and demonstrates the potential of VAEs for information compression in images. Beyond microscopy, the VAE-EM framework applies to inverse problems where simulated training data introduces a reality gap and where non-injective mappings would otherwise prevent unique solutions.
\end{abstract}


\section{Introduction}
\IEEEPARstart{E}{lectron} microscopy enables atomic-resolution imaging with applications across materials science, life sciences, and the semiconductor industry. Calibrating these microscopes requires frequent adjustments to compensate for optical aberrations that deteriorate image quality. Calibration is performed by analyzing special diagnostic images called Ronchigrams, which are diffraction patterns whose features reveal information about the microscope's current aberration state~\citep{Schnitzer2019}.
Currently, the calibration is typically performed or monitored by experts who interpret these patterns and determine appropriate corrections, making calibration expensive and time-consuming. For this reason, automating the calibration process is highly desirable.

The challenge in automating the calibration process comes from the complex, non-injective relationship between aberration parameters and the Ronchigram~\citep{VanHulst2025}. Different aberration states can produce indistinguishable diffraction patterns. This ambiguity, combined with the high dimensionality of Ronchigram images (typically around $10^5$ pixels) and the high sensitivity of the microscope to environmental factors, makes it difficult to determine the optimal calibration parameters.

While significant progress has been achieved over the past decade in the automation of STEM calibration, it is still an area of active research. Existing approaches broadly fall into two categories: physics-based and data-driven methods. Physics-based approaches were developed first, employing physical optics models to relate Ronchigram features to aberration parameters~\citep{CEOS,Tejada2009,Vulovic2012,Sherpa2019,Rudnaya2011}.
These methods are generally robust to different operating conditions, but are fundamentally limited by the accuracy of the underlying physical models and by the inability to adapt to day-to-day variations in real electron microscopes.

In contrast, data-driven calibration methods remove the reliance on detailed physics models (see~\citep{VanHulst2025} for a comprehensive review). These methods typically employ deep learning to produce a scalar quality metric for each image that can be optimized~\citep{Schubert2024,Ma2024a,Pattison2024,Bertoni2023}. Optimizing such a scalar value is straightforward, but has its limitations. The scalar goal rapidly loses information content as the number of calibration parameters increases. Additionally, the referenced data-driven methods are trained on simulated data (due to limited experimental data availability), creating a simulation-to-reality gap when deployed on real microscopes. 
Despite these limitations, data-driven methods remain more promising for achieving robust calibration across different microscopes and operating conditions, as they can adapt to day-to-day variations without requiring accurate physical models.

Although the data-driven calibration methods developed thus far have shown promising results, the reliance on scalar representations and simulation data fundamentally limits their performance and applicability. This paper aims to address these two limitations directly. First, we introduce multi-dimensional representations through variational autoencoders (VAEs)~\citep{Kingma2014}. Second, we adopt a simultaneous state and model estimation approach based on expectation maximization (EM)~\citep{Dempster1977} to explicitly account for the simulation-to-reality gap. This combination of VAE-based dimensionality reduction with EM-based joint estimation is, to our knowledge, novel in the context of inverse problems. Additionally, we introduce the use of symmetry-constrained kernels to establish global identifiability, which presents a theoretical contribution that addresses a fundamental limitation of standard EM approaches for inverse problems.
We model the map from calibration parameters to VAE latent representations using Gaussian processes (GPs)~\citep{Rasmussen2006}. GPs provide a flexible, non-parametric framework that can embed prior knowledge of the problem structure. The key insight that enables the joint state and model estimation approach is that the aberrations exhibit a known symmetrical structure~\citep{Scherzer1936}. By enforcing these symmetries in the GP model, we obtain global identifiability of the joint estimation problem, a property that is typically absent in standard EM approaches for inverse problems~\citep{Wu1983,Tarantola2005}. Additionally, we employ an adaptive refinement strategy for the discrete set of aberration states considered in the EM algorithm that improves estimation accuracy even when considering a small number of aberration state candidates. This refinement strategy takes inspiration from particle filtering methods~\citep{Arulampalam2002}. Finally, we develop an input selection strategy that chooses calibration inputs to maximize information gain to accelerate convergence.

Our contributions can be summarized as follows:
\begin{enumerate}
    \item We introduce variational autoencoders in electron microscopy to learn multi-dimensional latent representations of Ronchigram images. These representations capture more information than scalar quality metrics used in prior work.
    \item We develop a novel VAE-EM framework for joint state and model estimation that addresses the simulation-to-reality gap inherent in training on simulated data. We establish global identifiability of this joint estimation problem by exploiting symmetry constraints from Fourier optics.
    \item We validate on real STEM data, showing that the proposed method more than halves the estimation error of existing automated calibration methods and converges to accurate estimates within tens of seconds.
\end{enumerate}
While developed for STEM calibration, the methodological contributions are of general interest and extend beyond this application domain. The VAE-EM framework and identifiability analysis apply to inverse problems where known structural constraints (such as evenness) can be exploited to restore unique solutions in joint state-model estimation. We discuss the conditions for general applicability in Section~\ref{sec:discussion}.

The remainder of this paper is organized as follows. Section~\ref{sec:problem_formulation} provides the problem formulation, including the STEM imaging model and preprocessing steps. Section~\ref{sec:VAE} details the proposed VAE architecture and training procedure. Section~\ref{sec:EM} presents the EM algorithm for joint state and model estimation, along with identifiability analysis. Section~\ref{sec:results} showcases experimental results on real STEM data, and Section~\ref{sec:discussion} discusses the implications of our findings.

\section{Problem Formulation}
\label{sec:problem_formulation}
The electron microscope calibration problem and experimental setup are presented in detail in our previous work~\citep{VanHulst2025}. In short, the problem involves overcoming optical aberrations in scanning transmission electron microscopy (STEM), as these aberrations deteriorate image quality. This is achieved by interpreting high-dimensional Ronchigram images to determine the calibration parameters that result in the best image quality. In this section, we summarize the key aspects and structure of the calibration problem including the aberration model, Ronchigram observations and preprocessing steps.

\subsection{Aberration Optics and Ronchigrams}
\begin{figure*}[t]
    \centering
    \includegraphics[width=\textwidth]{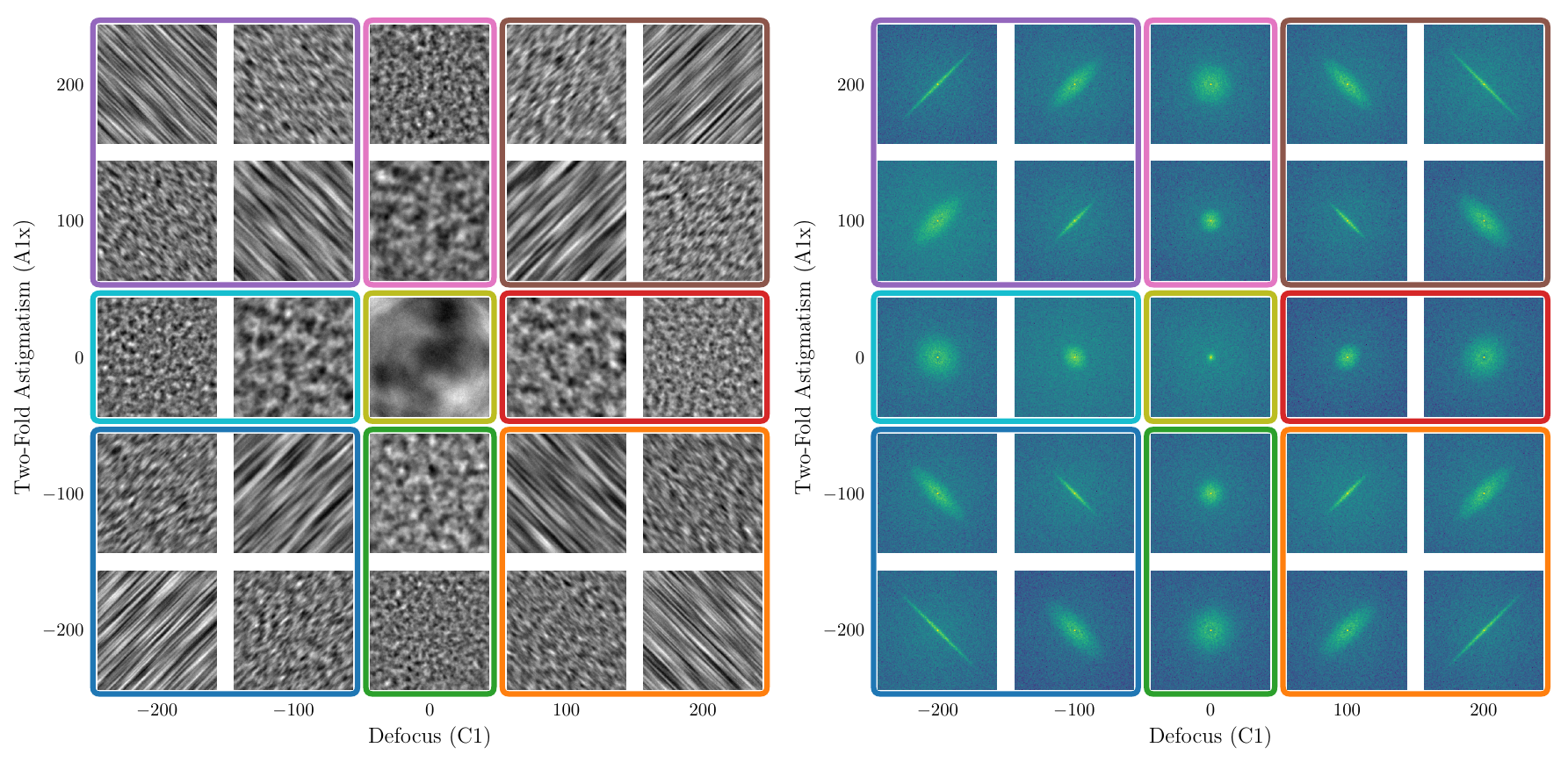}
    \caption{Grid of Ronchigram images (left) and their Fourier power spectra (right) for different aberration values. The horizontal axis varies defocus ($C1$) from $-200$ to $+200$~nm, while the vertical axis varies two-fold astigmatism ($A1x$) over the same range. The center images correspond to well-tuned aberrations ($C1=0$, $A1x=0$). Colored rectangles group images in specific aberration regions. Note that the Fourier power spectra exhibit symmetry about the origin, with images at $(x_1, x_2)$ and $(-x_1, -x_2)$ appearing similar (e.g., purple and orange, cyan and red).}
    \label{fig:images_grid}
\end{figure*}
Let the microscope aberration state $x(t) \in \mathbb{R}^n$ represent the aberration levels at discrete time $t \in \mathbb{N}$, where $x(t) = 0$ corresponds to perfect calibration. The aberrations evolve according to:
\begin{equation}
\label{eq:aberration_dynamics}
x(t+1) = x(t) + u(t),
\end{equation}
where $u(t) \in \mathbb{R}^n$ represents the user-selected calibration inputs applied to compensate aberrations. Over typical calibration timescales, aberrations remain constant unless nonzero inputs are applied. This implies that the evolution of the aberrations is deterministic for our purposes. Additionally, if the aberration state becomes known, we can immediately achieve optimal calibration by applying the input $u(t) = -x(t)$. To alleviate notation, we denote the cumulative input up to time $t$ as $s(t) = \sum_{\tau=0}^{t-1} u(\tau)$, so that $x(t) = x_0 + s(t)$.

The STEM produces high-dimensional Ronchigram observations $y(t) \in \mathbb{N}_0^p$ (where $\mathbb{N}_0 = \{0, 1, 2, \ldots\}$) representing electron counts at $p$ pixels (with $p$ typically in the order of tens of thousands). For a fixed sampling location and beam intensity, the observations follow a Poisson distribution. The rate of this distribution is then determined by the aberration state $x(t)$ through an unknown image formation function $g: \mathbb{R}^n \to \mathbb{R}_{+}^p$ (where $\mathbb{R}_+ = (0, \infty)$). We model
\begin{equation}
\label{eq:observation_model}
y(t) \sim \mathrm{Poisson}\left(g(x(t))\right),
\end{equation}
where the Poisson distribution applies element-wise to each pixel independently.

We employ two datasets of Ronchigram images in this work. The first is a large simulated dataset generated using a high-fidelity wave optics simulator based on the multi-slice method~\cite{Kirkland2010} that contains many different aberration values. This dataset is used for VAE training, which requires substantial data volumes. While the simulator is relatively accurate, it inherently suffers from undermodeling and does not capture day-to-day variations present in the experimental microscope. The second dataset consists of experimental Ronchigrams collected from a real STEM system on separate operating days. In both datasets, the aberration states are known since they are set by the user. These states can be used to validate our calibration procedure by comparing estimated aberrations to ground truth values. Importantly, the known aberration values in the simulated dataset also enable us to optimize hyperparameters such as the prior mean and covariance functions that will be used for Gaussian process regression in Section~\ref{sec:EM}.

\subsection{Preprocessing and symmetry of the Ronchigrams}
\label{sec:symmetry_derivation}
To extract useful information from the raw Ronchigram images $y(t)$, we perform several preprocessing steps. These preprocessing steps follow standard practices in automated STEM calibration and are laid out in detail in our previous work~\citep{VanHulst2025}. To summarize, the workflow consists of three operations applied to each image vector $y(t)$:
\begin{enumerate}
    \item Apply a Hann window element-wise to reduce spectral leakage.
    \item Transform to the 2D Fourier domain by treating the $p$-dimensional vector as an image with spatial pixel coordinates.
    \item Compute the power spectrum by taking the element-wise squared modulus.
\end{enumerate}
We denote the composite preprocessing operator by $\mathcal{P}$, so that the preprocessed observations are
\begin{equation}
\label{eq:preprocessed_observation}
	\tilde{y}(t) = \mathcal{P}\{y(t)\},
\end{equation}
where $\mathcal{P}: \mathbb{N}_0^p \to \mathbb{R}_+^p$ includes windowing, the 2D discrete Fourier transform, and the squared modulus operation. The technical definition of the 2D Fourier transform in terms of spatial coordinates and frequency indices is provided in the supplementary material. After preprocessing, the observations $\tilde{y}$ can be modeled as a noisy function of the aberrations
\begin{equation}
\label{eq:preprocessed_observation_model}
	\tilde{y}(t) = h(x(t)) + \eta(t),
\end{equation}
where $h(x) := \mathbb{E}[\mathcal{P}\{y\} \mid x]$ denotes the expected power spectrum as a function of the aberration state and $\eta(t) := \tilde{y}(t) - h(x(t))$ captures the measurement noise. The distribution of $\eta$ is typically non-trivial due to the nonlinear preprocessing steps and the Poisson noise, but we do not require specific distributional assumptions for $\eta$ at this point. Fig.~\ref{fig:images_grid} shows example Ronchigram images and their preprocessed Fourier power spectra for different aberration values. A key result of the preprocessing is that $h$ is even with respect to the aberrations:
\begin{equation}
\label{eq:symmetry_property}
h(-x) = h(x) \quad \forall x \in \mathbb{R}^n.
\end{equation}
This symmetry arises from the physics of electron optics and appears only in the Fourier power spectrum, not in the spatial domain (thus, in general, $g(-x) \neq g(x)$). The full derivation of this property from first principles is provided in the supplementary material.
This derivation uses continuous Fourier analysis in the spatial frequency domain to establish the symmetry for the continuous intensity function, then shows that the discrete 2D Fourier transform preserves this symmetry after sampling onto a pixel grid. The symmetry property is central to our approach, as it enables unique identifiability in the joint estimation problem developed in Section~\ref{sec:EM}.

\subsection{Approach}
\label{subsec:problem_formulation}
\begin{figure*}[t]
    \centering
    \includegraphics[width=0.98\textwidth]{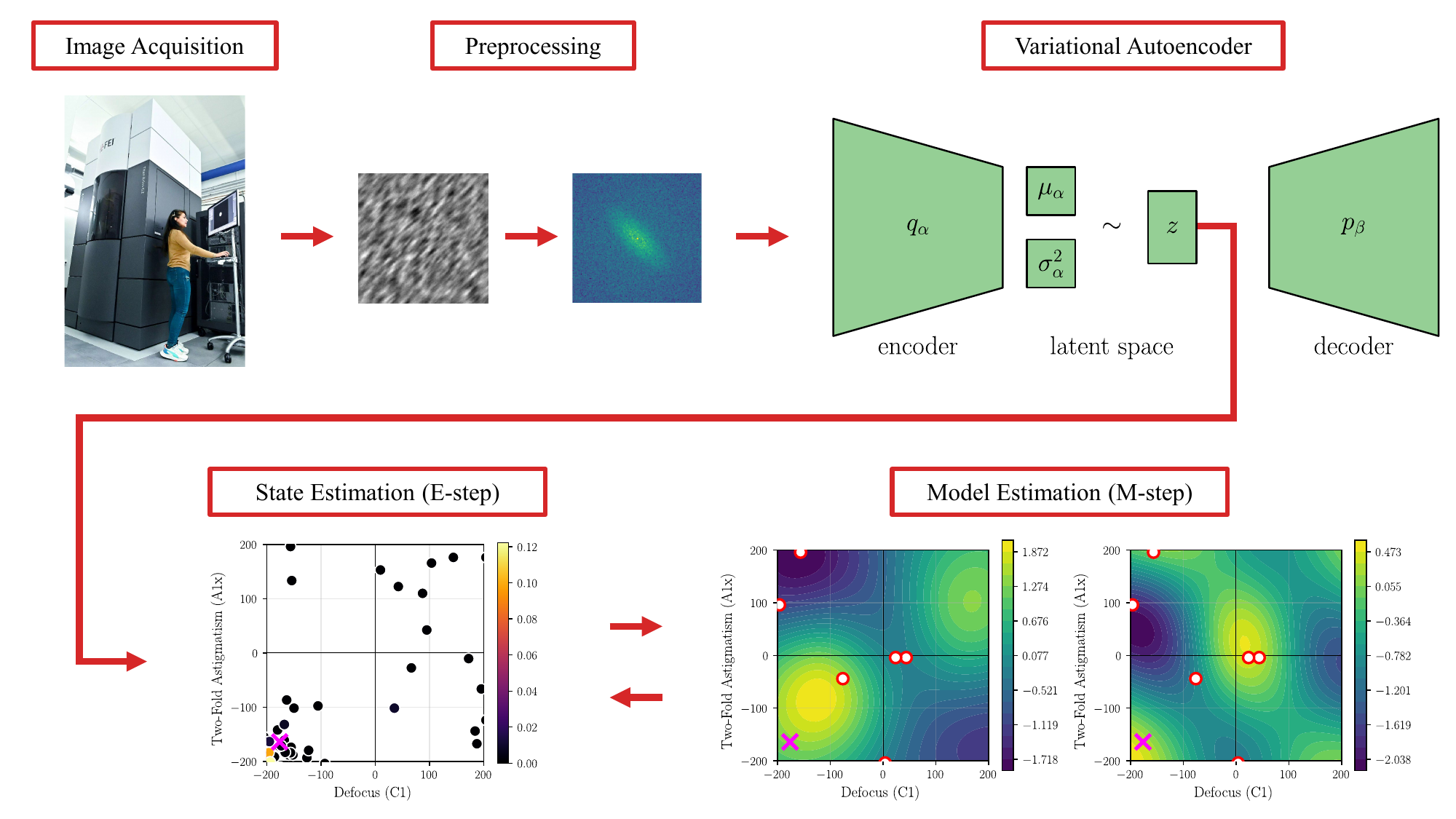}
    \caption{Calibration pipeline overview. High-dimensional Ronchigram images are preprocessed and then encoded to a low-dimensional latent space using a VAE. The EM algorithm then iteratively refines both the aberration state estimate and the mapping from aberrations to latent representations. In the E-step, the distribution over aberration states is estimated by evaluating the likelihood of observed latent space values for different candidate states. These candidates (\candidatemarkerlow\candidatemarkermid\candidatemarkerhigh{}) are highlighted in the aberration space, where the color indicates their estimated likelihood (low \candidatemarkerlow{} to high \candidatemarkerhigh). The \groundtruthmarker{} indicates the ground truth aberration state. Note that this state and its symmetrical neighbor have an increased density of candidates due to the adaptive candidate refinement strategy. The most likely candidates are also clustered around the ground truth state. The M-step then updates the Gaussian process model of the aberration to latent space mapping. The current model belief based on 5 data points (\datamarker) is illustrated as a function of the aberrations. Note that this model belief is symmetric due to the symmetry constraint in the model structure. The symmetry point is not necessarily at the origin (in the true aberration coordinates) due to the fact that the state estimate is not yet perfect. The M- and E-steps are iterated until convergence.}
    \label{fig:framework_overview}
\end{figure*}
Our goal is to calibrate the STEM by making the aberration state $x$ zero. This goal can be equivalently characterized as the estimation of the aberration state $x(t)$ from the observations $\{y(t)\}_{t=0}^T$ for corresponding chosen inputs $\{u(t)\}_{t=0}^{T-1}$, since we have full knowledge of the dynamics and control inputs. Additionally, estimating the full trajectory $\{x(t)\}_{t=0}^T$ is equivalent to estimating the initial state $x_0 = x(0)$ that then determines all future states via the deterministic dynamics $x(t) = x_0 + s(t)$.

The estimation problem is challenging for two fundamental reasons. First, the Ronchigram observations $y(t) \in \mathbb{R}_+^p$ are extremely high-dimensional (typically $p \approx 10^5$ pixels), making direct estimation computationally intractable and statistically inefficient. Second, the image formation function $g: \mathbb{R}^n \to \mathbb{R}_+^p$ (and similarly $h: \mathbb{R}^n \to \mathbb{R}_+^p$) is unknown and varies day-to-day due to unmodeled effects, environmental conditions, and instrument drift. While physics-based models exist, they are insufficiently accurate for precise and adaptive calibration. Additionally, the choice of inputs $\{u(t)\}_{t=0}^{T-1}$ can significantly affect the estimation convergence rate, so input selection is an important consideration that we address.

To address these challenges, we propose a two-stage approach:
\begin{enumerate}
    \item \textbf{Dimensionality reduction (Section~\ref{sec:VAE}):} Use a variational autoencoder (VAE) to learn a low-dimensional latent representation $z(t) \in \mathbb{R}^\ell$ (where $\ell \ll p$) of the preprocessed observations $\tilde{y}(t)$ that preserves as much information about $x(t)$ as possible. As will be shown in the next section, this results in a simplified observation model $z(t) = f(x(t)) + \nu(t)$ where $f: \mathbb{R}^n \to \mathbb{R}^\ell$ is the effective latent mapping, and $\nu(t)$ is the noise whose characteristics are discussed in Section~\ref{sec:VAE}.
    \item \textbf{Joint state and model estimation (Section~\ref{sec:EM}):} Given the latent observations and chosen inputs $\{z(t)\}_{t=0}^{T}, \{u(t)\}_{t=0}^{T-1}$, simultaneously estimate both the initial state $x_0$ and the latent mapping $f$ using an Expectation-Maximization (EM) algorithm with provable identifiability guarantees.
\end{enumerate}
The full calibration pipeline proposed in this paper is illustrated in Fig.~\ref{fig:framework_overview}. In our previous work, we also proposed a two-stage approach consisting of Ronchigram interpretation and decision-making based on the interpretations~\citep{VanHulst2025}. However, this approach relied on scalar image representations and had no explicit model estimation step. 

\section{Learning Compact Image Representations with VAEs}
\label{sec:VAE}
We employ a variational autoencoder (VAE) to learn a low-dimensional latent representation $z \in \mathbb{R}^\ell$ (where $\ell \ll p$) of the preprocessed Ronchigram power spectra $\tilde{y}$.

\subsection{VAE Architecture}
\begin{figure}[t]
    \centering
    \begin{minipage}{0.06\textwidth}
        \centering
        \includegraphics[width=\textwidth]{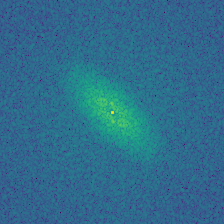}
        \end{minipage}
    \hspace{0\textwidth}
    \begin{minipage}{0.3\textwidth}
        \centering
        \includegraphics[width=\textwidth]{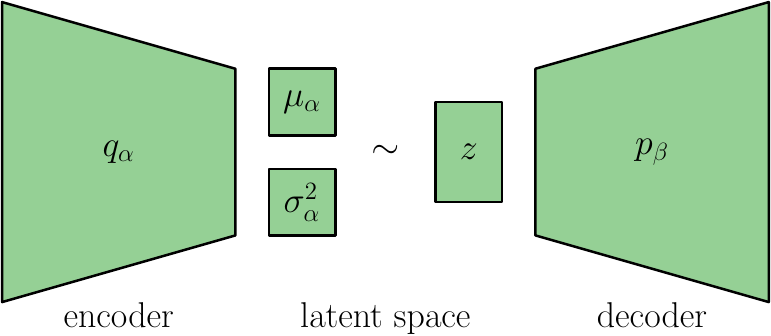}
    \end{minipage}
    \hspace{0\textwidth}
    \begin{minipage}{0.06\textwidth}
        \centering
        \includegraphics[width=\textwidth]{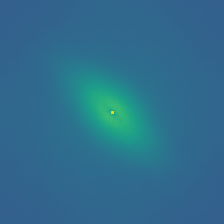}
    \end{minipage}
    \caption{Variational Autoencoder architecture. The encoder maps preprocessed Ronchigram power spectra to a low-dimensional latent distribution, from which samples are decoded to reconstruct the input. The weights and biases of the encoder and decoder networks are optimized to maximize the ELBO objective in~\eqref{eq:elbo}, which balances reconstruction accuracy and latent space regularization. The example reconstruction to the right of the decoder resembles the input and shows the denoising effect of the VAE.}
    \label{fig:vae_architecture}
\end{figure}

The VAE consists of an encoder network $q_\alpha(z|\tilde{y})$ and decoder network $p_\beta(\tilde{y}|z)$
\begin{equation}
\label{eq:vae_encoder}
\begin{aligned}
    q_\alpha(z|\tilde{y}) &= \mathcal{N}(z; \mu_\alpha(\tilde{y}), \sigma^2_\alpha(\tilde{y}) I_\ell), \\
    p_\beta(\tilde{y}|z) &= \mathcal{N}(\tilde{y}; \mu_\beta(z), \sigma^2_\beta I_p),
\end{aligned}
\end{equation}
where $\alpha$ and $\beta$ are the encoder and decoder parameters, respectively. The encoder maps the input $\tilde{y}$ to a Gaussian distribution in the latent space with mean $\mu_\alpha(\tilde{y})$ and diagonal covariance $\sigma^2_\alpha(\tilde{y}) I_\ell$. The decoder reconstructs the input from latent samples $z$ using a Gaussian likelihood with mean $\mu_\beta(z)$ and fixed variance $\sigma^2_\beta I_p$. See Fig.~\ref{fig:vae_architecture} for a visualization of the VAE architecture.

We implement both encoder and decoder using convolutional neural networks (CNNs) to process the Ronchigram images efficiently. The encoder consists of five convolutional layers with progressively increasing channel dimensions $[16, 32, 64, 128, 256]$, each followed by batch normalization and LeakyReLU activation (negative slope $0.01$). Strided convolutions (stride 2) perform downsampling. The final convolutional output is flattened and passed through fully connected layers to produce the latent mean $\mu_\alpha(\tilde{y})$ and log-variance $\log \sigma^2_\alpha(\tilde{y})$. The decoder mirrors this architecture with transposed convolutions for upsampling. We set the latent dimension $\ell = n$ to match the number of aberration parameters. The symmetry constraint $h(-x) = h(x)$ means the encoder must map both $x$ and $-x$ to the same latent representation, but this identification of antipodal points does not reduce the intrinsic dimensionality of the parameter space. Formally, the quotient space $\mathbb{R}^n / \sim$ (where $x \sim -x$) remains $n$-dimensional, so $\ell = n$ suffices to represent all distinguishable aberration states. In practice, one might choose $\ell > n$ to allow the capturing of additional variations in the data, but in our setting we were able to mitigate such variations through careful data generation and preprocessing.

The parameters of the VAE $\alpha$ and $\beta$ are optimized to maximize the evidence lower bound (ELBO)
\begin{equation}
\label{eq:elbo}
\mathcal{L} = \mathbb{E}_{q_\alpha(z|\tilde{y})}[\log p_\beta(\tilde{y}|z)] - \text{KL}(q_\alpha(z|\tilde{y}) \| p(z)),
\end{equation}
where $p(z) = \mathcal{N}(0, I_\ell)$ is the prior distribution and $\text{KL}$ denotes Kullback-Leibler divergence. The first term encourages accurate reconstruction of the input despite the low-dimensional latent space, while the second term regularizes the latent distribution to be close to the prior, which is typically chosen as a standard normal distribution. This regularization encourages smoothness and structure in the latent space and typically improves generalization~\citep{Kingma2014}.

\subsection{Latent-Space Observation Model}
\begin{figure}
    \centering
    \includegraphics[width=0.45\textwidth]{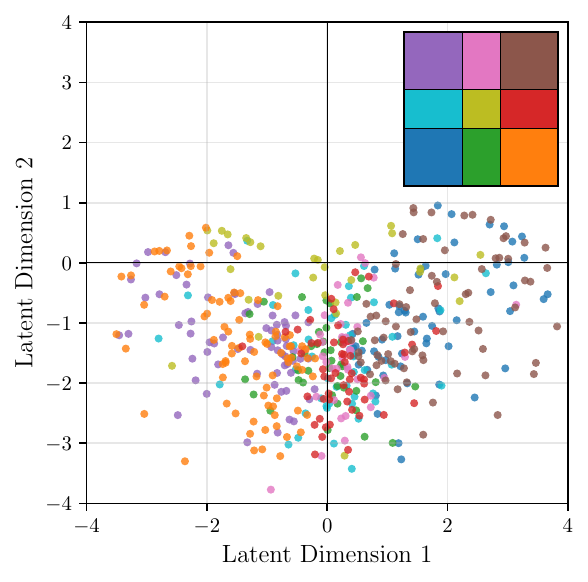}
    \caption{Learned VAE latent space for simulated data that contains two varying aberration parameters: defocus ($C1$), and two-fold astigmatism in the x-direction ($A1x$). The color of the data points is determined by the aberration region, with the assignment shown in the top right of the figure and in Fig.~\ref{fig:images_grid}. We can observe that similar aberrations are grouped together. Additionally, different aberration values that produce similar Ronchigrams are still mapped to the same latent space values (example: purple and orange), demonstrating the non-injectivity of the observation model.}
    \label{fig:vae_latent_space_sim}
\end{figure}

The result of VAE training can be seen in Fig.~\ref{fig:vae_latent_space_sim}, which shows a latent space learned from simulated Ronchigrams with two varying aberration parameters. We limit the number of varying aberrations and latent space dimensions to two for visualization purposes. By color-coding the latent representations according to aberration values, we observe that the VAE groups similar aberration values together in the latent space. Additionally, distinct aberration values that produce visually similar Ronchigrams are also mapped to similar latent representations, reinforcing the non-injectivity of the observation model. Finally, we can observe the evenness property in the latent space: points with opposite aberration values ($x$ and $-x$) are mapped to the same latent representation.

After training, the encoder mean $\mu_\alpha(\tilde{y})$ provides the latent representation. This establishes the latent-space observation model that will be used for state estimation:
\begin{equation}
    \begin{aligned}
        x(t+1) &= x(t) + u(t) \\
        z(t) &= f(x(t)) + \nu(t)
    \end{aligned}
\label{eq:latent_observation_model}
\end{equation}
where $f: \mathbb{R}^n \to \mathbb{R}^\ell$ is the composition of the encoder mean $\mu_\alpha$ and $h$ from~\eqref{eq:preprocessed_observation}, and $\nu(t)$ denotes the stochastic variation in the latent space induced by the Poisson noise in~\eqref{eq:observation_model}. The latent observations $z(t) \in \mathbb{R}^\ell$ have much lower dimension than the original Ronchigrams ($\ell \ll p$).

Crucially, because the VAE is trained on data satisfying $h(-x) = h(x)$ (established in Section~\ref{sec:symmetry_derivation}), the learned latent mapping $f = \mu_\alpha \circ h$ inherits this symmetry by composition: $f(-x) = \mu_\alpha(h(-x)) = \mu_\alpha(h(x)) = f(x)$ for all $x \in \mathbb{R}^n$.

We assume $\nu(t) \sim \mathcal{N}(0, \Sigma_\nu)$, with $\Sigma_\nu \succ 0$, is i.i.d. Gaussian noise independent of $x(t)$. This is justified in our setting by the delta-method and central limit theorem heuristics: under mild smoothness of $\mu_\alpha$ and when each latent coordinate aggregates many independent pixel contributions, different Poisson realizations of $y(t)$ induce approximately Gaussian fluctuations in $\mu_{\alpha}(\mathcal{P}\{y(t)\})$~\citep{Nilsen2022,Sirignano2020,Monchot2023}.

With the low-dimensional latent representations $z(t)$ in place, estimating $x_0$ from $\{z(t)\}_{t=0}^{T}, \{u(t)\}_{t=0}^{T-1}$ becomes more tractable. If the VAE is trained on representative real data, the latent mapping $f$ can be estimated directly from training data and standard state estimation methods (such as maximum likelihood or Bayesian filtering) can be applied to estimate $x_0$. This constitutes a VAE-based state estimation approach that is of independent interest. However, when the VAE is trained on simulated data, the simulation-to-reality gap means that $f$ cannot be assumed known; it must also be estimated to account for undermodeling and day-to-day differences in the STEM. The next section develops a joint estimation framework to simultaneously estimate both $x_0$ and $f$ while exploiting the symmetry properties established in Section~\ref{sec:symmetry_derivation}. Prior knowledge about $f$ from the simulated dataset is leveraged through Bayesian priors to accelerate convergence, but the framework does not rely on it being exact. As a special case, this framework recovers the VAE-based state estimation approach when $f$ is known from training data. The framework also enables adaptive input selection to accelerate convergence.

\section{Joint State and Model Estimation using Expectation Maximization}
\label{sec:EM}
This section establishes that the joint estimation of $x_0$ and $f$ is globally identifiable for symmetrical observation models. We then detail an Expectation-Maximization (EM) algorithm to solve the joint estimation problem.

\subsection{Identifiability of the joint estimation problem}
\label{sec:identifiability}
The joint estimation of both the initial state $x_0$ and the latent mapping $f$ from $\{z(t)\}_{t=0}^{T}, \{u(t)\}_{t=0}^{T-1}$ is fundamentally ambiguous due to translation symmetry and bilinear coupling. This section establishes conditions under which global identifiability is guaranteed. Throughout this section, we assume a linear basis function representation $f(x) = \Phi(x)^\top \theta$ where $\Phi(x) = [\phi_1(x), \dots, \phi_m(x)]^\top$ is a vector of basis functions and $\theta \in \mathbb{R}^m$ are parameters. In practice, the results from this section extend to Gaussian process models (Section~\ref{sec:EM}), which for a given dataset can be interpreted as a linear model in an infinite-dimensional feature space induced by the kernel~\citep{Rasmussen2006}.

In this setting, the joint least squares problem
\begin{equation}
\label{eq:joint_estimation}
(x_0^*, \theta^*) = \arg\min_{x_0, \theta} \sum_{t=0}^{T-1} \|z(t) - \Phi(x_0 + s(t))^\top \theta\|^2
\end{equation}
is fundamentally ambiguous due to the coupling between $x_0$ and $\theta$. The following lemma formalizes this ambiguity.

\begin{lemma}
    \label{lem:ambiguity}
    Let $\Phi: \mathbb{R}^n \to \mathbb{R}^m$ be a vector of basis functions, and $(x_0^*, \theta^*)$ be a solution to~\eqref{eq:joint_estimation}. If
    \begin{equation}
        \label{eq:translation_invariance}
        \Phi(x)^\top \theta^* = \Phi(x + \delta)^\top \bar{\theta} \quad \forall x \in \mathbb{R}^n
    \end{equation}
    holds for some nonzero shift $\delta \in \mathbb{R}^n$, $\delta \neq 0$, and vector $\bar{\theta} \in \mathbb{R}^m$, then the pair $(x_0^* + \delta, \bar{\theta})$ is also a solution, making the optimization problem~\eqref{eq:joint_estimation} ill-posed.
\end{lemma}
The proof follows from direct substitution (see the supplementary material). To recover the identifiability of the unknown parameters and make the optimization problem~\eqref{eq:joint_estimation} well-posed, we impose structural and spectral conditions on the basis functions. To analyze these conditions, we employ the continuous Fourier transform in the aberration parameter space $\mathbb{R}^n$ (note: this is distinct from the 2D image-space Fourier transform described in the supplementary material). For a function $\phi: \mathbb{R}^n \to \mathbb{R}$ mapping aberration states to scalars, the Fourier transform is defined by
\begin{equation}
\label{eq:continuous_fourier}
\widehat{\phi}(\omega) = \int_{\mathbb{R}^n} \phi(x) e^{-i\omega^\top x} dx,
\end{equation}
where $\omega \in \mathbb{R}^n$ is the frequency variable in aberration space. We impose the following conditions on the basis functions $\{\phi_i\}_{i=1}^m$ and the true parameter vector $\theta \in \mathbb{R}^m$:
\begin{enumerate}[label=(C\arabic*)]
    \item \textbf{Evenness.} Each basis function $\phi_i$ is even: $\phi_i(-x) = \phi_i(x)$ for all $x \in \mathbb{R}^n$. Consequently, $\widehat{\phi}_i(\omega)$ is real-valued and even for all $\omega \in \mathbb{R}^n$.
    \item \textbf{Continuous spectral support.} For any nonzero $c \in \mathbb{R}^m$, the spectral amplitude $\widehat{\Phi}(\omega)^\top c$ (where $\widehat{\Phi}(\omega) = [\widehat{\phi}_1(\omega), \ldots, \widehat{\phi}_m(\omega)]^\top$) is not concentrated on a discrete set. That is, no nontrivial linear combination of basis functions is periodic.
    \item \textbf{Spectral richness.} There exists a set $\Omega \subset \mathbb{R}^n$ with positive Lebesgue measure such that the set of vectors $\{\widehat{\Phi}(\omega) : \omega \in \Omega\} \subset \mathbb{C}^m$ spans $\mathbb{C}^m$. 
    \item \textbf{Nontrivial signal.} For the optimal parameter vector $\theta^* \in \mathbb{R}^m$, the spectral amplitude $S(\omega) := \widehat{\Phi}(\omega)^\top \theta^*$ is nonzero on a set $\Omega_0 \subseteq \mathbb{R}^n$ with positive Lebesgue measure.
\end{enumerate}
Note that condition (C2) automatically implies that the basis functions are linearly independent for any fixed $x_0$, since linearly dependent bases would allow construction of $c$ such that the spectral amplitude is zero everywhere (the empty set being discrete). Condition (C3) ensures that the parameter vector $\theta$ can be uniquely recovered from the spectral representation $\widehat{\Phi}(\omega)^\top \theta$ observed over $\Omega$. Condition (C4) ensures that the true mapping $f(x) = \Phi(x)^\top \theta$ is not identically zero, as that would make estimation of $x_0$ impossible. Conditions (C3) and (C4) are more technical and less intuitive, but are naturally satisfied by common function classes. Conditions (C1)--(C4) together guarantee global identifiability, as formalized in the following theorem.

\begin{theorem}[Global Identifiability of $(x_0, \theta)$]\label{thm:identifiability}
Let the basis functions $\{\phi_i\}_{i=1}^m$ and the true parameter vector $\theta^* \in \mathbb{R}^m$ satisfy conditions (C1)--(C4). If two parametrizations produce identical measurement functions:
\begin{equation}\label{eq:function_equivalence}
\Phi(x + \xi)^\top \theta = \Phi(x + \xi')^\top \theta' \quad \forall x \in \mathbb{R}^n,
\end{equation}
then necessarily $\xi = \xi'$ and $\theta = \theta'$. Consequently, the joint optimization problem~\eqref{eq:joint_estimation} has a unique global minimizer, and the pair $(x_0, f)$ is globally identifiable.
\end{theorem}

\begin{remark}
The assumption that $f$ lies exactly in the span of $\Phi$ can be relaxed. If $f$ is well-approximated by its projection onto the basis, i.e., $\|f(x) - \Phi(x)^\top \theta^*\| \leq \epsilon$ uniformly for some $\epsilon > 0$, then the identifiability result extends to the projected function. The key insight is that translation ambiguity (Lemma~\ref{lem:ambiguity}) exists regardless of whether $f$ lies exactly in the span or is merely well-approximated. Under conditions ensuring uniform convergence of the approximation as the basis expands (which are typically satisfied for GP models with universal kernels and well-behaved functions), conditions (C1)--(C4) continue to guarantee unique recovery of $x_0$ up to approximation error $O(\epsilon)$.
\end{remark}

The proof is given in the supplementary material. In the next section, we detail a numerical approach to jointly estimate $x_0$ and $f$ that leverages these identifiability results.

\subsection{EM algorithm for joint state and model estimation}
\label{subsec:EM_algorithm}
Under conditions (C1)--(C4) established in Theorem~\ref{thm:identifiability}, we can jointly estimate the initial aberration state $x_0$ and the latent mapping $f$. We do so using an Expectation-Maximization (EM) approach. We model $f$ using a Gaussian process (GP) to generate flexible, non-parametric representations that also provide uncertainty quantification. We discretize the initial state space into a finite set of candidate states to enable tractable computation of the E-step. Details on the selection and refinement of these candidate states are provided in Section~\ref{subsec:EM_implementation}.

Unlike standard EM applications where the M-step (model parameter estimation) is the primary goal, our setting emphasizes the E-step, as it directly yields the aberration state estimate $x_0$. This focus on state estimation is motivated by the calibration objective and enabled by the identifiability results from Section~\ref{sec:identifiability}.

\subsubsection{Gaussian process representation}
We model the latent mapping $f$ using a Gaussian process (GP) prior. This modeling choice provides flexibility and computational tractability. The GP framework can approximate a broad class of functions arbitrarily well with appropriate kernel selection~\citep{Rasmussen2006} and allows us to quantify uncertainty in the learned mapping. While our choice of the squared exponential kernel (detailed below) imposes smoothness assumptions on $f$, this is physically motivated: we expect the aberration-to-latent mapping to vary continuously with aberration parameters. The framework readily accommodates alternative kernels (e.g., Mat\'ern, rational quadratic) if different smoothness or regularity properties are desired.

We assume a Gaussian process prior on $f$:
\begin{equation}
\label{eq:gp_prior}
    f \sim \mathcal{GP}(\mu_f, k(\cdot, \cdot)),
\end{equation}
with mean function $\mu_f: \mathbb{R}^n \to \mathbb{R}$ and covariance (kernel) function $k: \mathbb{R}^n \times \mathbb{R}^n \to \mathbb{R}$. For notational simplicity, we consider the scalar case ($\ell = 1$) in the equations presented in this section. The extension to multi-dimensional latent spaces ($\ell > 1$) is straightforward. Each output dimension can be modeled independently using a separate GP with the same kernel structure. That is, $f(x) = [f_1(x), \ldots, f_\ell(x)]^\top$ where each $f_j \sim \mathcal{GP}(\mu_{f,j}, k(\cdot, \cdot))$. As a sidenote, we recover the parametric linear model with the basis $\Phi(x)$ from Section~\ref{sec:identifiability} as a special case of the GP framework using the kernel trick.

To enforce the symmetry condition (C1), we use a symmetrized kernel resulting from a standard non-symmetric one
\begin{equation}
\label{eq:symmetric_kernel}
    k(x, x') = \frac{1}{2}\left(k_0(x, x') + k_0(x, -x')\right),
\end{equation}
where $k_0$ is a (stationary) kernel, e.g., a multivariate squared exponential,
\begin{equation}
    k_0(x, x') = \sigma_f^2 \exp\left(-\frac{1}{2} (x - x')^\top L^{-1} (x - x')\right),
\end{equation}
with signal variance $\sigma_f^2 > 0$ and lengthscale matrix $L \succ 0$. This construction ensures that any function drawn from $\mathcal{GP}(\mu_f, k)$ satisfies $f(-x) = f(x)$ if $\mu_f$ is an even function. The satisfaction of the conditions (C2)--(C4) for the basis functions generated by such a kernel function is derived in the supplementary material.

In practice, the prior mean $\mu_f$ and kernel hyperparameters (such as $\Sigma_\nu$, $\sigma_f^2$ and $L$) can be estimated from the large simulated dataset using standard maximum likelihood or cross-validation methods~\citep{Rasmussen2006}. This allows us to incorporate prior knowledge from physics-based simulations while retaining the flexibility to adapt to day-to-day variations by fitting the GP posterior to the current observations.

\subsubsection{Candidate initial states and likelihood}
To estimate $x_0$ using the EM algorithm, we discretize the initial state space into a finite set of candidate states. This discretization enables tractable computation of the E-step by evaluating the likelihood of each candidate under the GP model. Consider a discrete set of candidate initial states $\{x_0^{(i)}\}_{i=1}^{N}$. The selection of the candidate set is discussed in Section~\ref{subsec:EM_implementation}. We place a uniform prior $p(x_0^{(i)}) = 1/N$ on the candidates. Given candidate $x_0^{(i)}$ and the deterministic dynamics~\eqref{eq:aberration_dynamics}, the aberration trajectory is $x(t) = x_0^{(i)} + s(t)$ for $t = 0, 1, \ldots, T$.

Under the GP prior~\eqref{eq:gp_prior} and the observation model~\eqref{eq:latent_observation_model} conditioned on candidate $x_0^{(i)}$, the latent observations have marginal distribution (for the scalar case $\ell = 1$):
\begin{equation}
\label{eq:gp_marginal}
p(Z | x_0^{(i)}) = \mathcal{N}(Z; \mu_Z^{(i)}, K^{(i)} + \Sigma_\nu I_{T+1}),
\end{equation}
where
\begin{equation}
Z = \begin{bmatrix} z(0), z(1), \ldots, z(T) \end{bmatrix}^\top \in \mathbb{R}^{T+1}
\end{equation}
is the stacked latent observation vector over the time horizon. The stacked prior mean vector $\mu_Z^{(i)} \in \mathbb{R}^{T+1}$ evaluates the prior mean function $\mu_f$ at the aberration trajectory implied by candidate $i$:
\begin{equation}
\mu_Z^{(i)} = \begin{bmatrix} \mu_f(x_0^{(i)} + s(0)) \\ \mu_f(x_0^{(i)} + s(1)) \\ \vdots \\ \mu_f(x_0^{(i)} + s(T)) \end{bmatrix}.
\end{equation}
The prior covariance matrix $K^{(i)} \in \mathbb{R}^{(T+1) \times (T+1)}$ has elements
\begin{equation}
K^{(i)}_{ts} = k(x_0^{(i)} + s(t), x_0^{(i)} + s(s)), \quad t, s = 0, 1, \ldots, T.
\end{equation}
Finally, $\Sigma_\nu > 0$ is a scalar (in the $\ell=1$ case) representing the variance of the latent observation noise $\nu(t)$ from~\eqref{eq:latent_observation_model}. The total covariance in~\eqref{eq:gp_marginal} combines the GP prior covariance $K^{(i)}$ with the observation noise $\Sigma_\nu I_{T+1}$. For $\ell > 1$, we stack all latent dimensions and extend the covariance structure using Kronecker products. All subsequent expressions extend naturally with this block structure. With these definitions, we can evaluate the GP marginal likelihood $p(Z | x_0^{(i)})$ for each candidate, which forms the basis for the EM algorithm detailed next.

\subsubsection{Evidence Lower Bound and EM iterations}
The EM algorithm finds $f$ and $x_0$ by maximizing the Evidence Lower Bound (ELBO) and alternating between estimating the posterior distribution over $x_0$ (E-step) and updating the (GP) representation of $f$ (M-step)~\citep{Dempster1977}. Let $\pi: \{x_0^{(1)}, \ldots, x_0^{(N)}\} \to [0,1]$ be a probability distribution over the candidate set, where $\pi(x_0^{(i)})$ denotes the probability assigned to candidate $i$ and $\sum_{i=1}^N \pi(x_0^{(i)}) = 1$. The ELBO as a function of $f$ and $\pi$ is given by
\begin{equation}
\label{eq:elbo_em}
\text{ELBO}(\pi, f) = \sum_{i=1}^{N} \pi(x_0^{(i)}) \log p(Z | x_0^{(i)}) - \text{KL}(\pi(x_0) \| p(x_0)),
\end{equation}
where $\text{KL}(\pi \| p) = \sum_{i=1}^N \pi(x_0^{(i)}) \log \frac{\pi(x_0^{(i)})}{p(x_0^{(i)})}$ is the standard discrete KL divergence. The two steps of the EM algorithm are detailed next.

\emph{M-step:} Update the GP representation of $f$ given the observations $Z$ and the current distribution $\pi$ over candidates. Since the ELBO's dependence on $f$ enters through the likelihoods $\log p(Z | x_0^{(i)})$, and these are maximized by the GP posterior conditioned on each candidate, we compute the GP posterior for each candidate $i$. For each candidate $x_0^{(i)}$, the GP posterior mean conditioned on that candidate and on the observations $Z$ is given by
\begin{equation}
\label{eq:single_gp_posterior}
\mu_{f|Z}^{(i)}(x) = \mu_f(x) + k_x^{(i)} (K^{(i)} + \Sigma_\nu I_{T+1})^{-1} (Z - \mu_Z^{(i)}),
\end{equation}
where $k_x^{(i)} = [k(x, x_0^{(i)}), k(x, x_0^{(i)} + s(0)), \ldots, k(x, x_0^{(i)} + s(T))]$ is the kernel slice evaluated at the query point $x$ and the trajectory points for candidate $i$. We can then form the overall posterior of $f$ as a mixture of these individual posteriors weighted by the distribution $\pi$ over candidates. We obtain
\begin{equation}
\label{eq:exma_m_step}
\mu_{f|Z}(x) = \sum_{i=1}^{N} \pi(x_0^{(i)}) \mu_{f|Z}^{(i)}(x),
\end{equation}
assuming $\ell = 1$. For $\ell > 1$, the expressions extend in a straightforward manner with the block structure.

\emph{E-step:} Update the distribution $\pi$ over candidates given the current GP representation of $f$. Maximizing the ELBO over $\pi$ subject to $\sum_{i=1}^N \pi(x_0^{(i)}) = 1$ yields $\pi(x_0^{(i)}) \propto p(x_0^{(i)}) p(Z | x_0^{(i)})$. Assuming a uniform prior over candidates $p(x_0^{(i)}) = 1/N$, we obtain the posterior $\pi$ by evaluating the likelihoods $p(Z | x_0^{(i)})$ for each candidate and normalizing. The log-likelihood of observing $Z$ given candidate $i$ is
\begin{equation}
\label{eq:log_likelihood_gp_explicit}
\begin{aligned}
\log p(Z | x_0^{(i)}) &= -\frac{1}{2} (Z - \mu_Z^{(i)})^\top (K^{(i)} + \Sigma_\nu I_{T+1})^{-1} (Z - \mu_Z^{(i)}) \\
&\quad - \frac{1}{2} \log |K^{(i)} + \Sigma_\nu I_{T+1}| - \frac{T+1}{2} \log(2\pi).
\end{aligned}
\end{equation}
We then compute the posterior weights as $w_i \propto p(Z | x_0^{(i)})$ and normalize such that $\sum_{i=1}^N w_i = 1$ to obtain the discrete distribution $\pi(x_0^{(i)}) = w_i$. For $\ell > 1$, the expressions extend with the block structure detailed earlier.

The EM algorithm is guaranteed to yield monotonic improvement of the ELBO, ensuring convergence to a local optimum~\citep{Dempster1977,Wu1983}. Theorem~\ref{thm:identifiability} establishes that under conditions (C1)--(C4), the pair $(x_0, f)$ is globally identifiable, meaning there exists a unique solution to the joint estimation problem. However, the identifiability result does not guarantee that the EM algorithm will converge to the global optimum. Establishing global convergence remains an open theoretical question, though empirically the algorithm converges reliably to accurate estimates in our experiments. 

An important observation about our particular setting is that the E-step does not depend on the full GP mixture posterior from the M-step, but only on the individual candidate likelihoods computed from the individual GP posteriors. With a uniform prior, the E-step weights remain constant across EM iterations for a fixed dataset. This means we can compute the E-step once per dataset, after which a single M-step produces the final estimates. In practice, we often run only a single EM iteration per dataset, significantly reducing computational cost.

An alternative to the EM approach presented above is \emph{hard EM} that replaces the full posterior distribution $\pi(x_0)$ with a point estimate at each iteration. Specifically, in the hard E-step, instead of computing the full distribution over candidates, we select $\hat{x}_0^{(i^*)}$ where $i^* = \arg\max_i \bar{w}_i$, effectively setting $\pi(x_0^{(i^*)}) = 1$ and $\pi(x_0^{(j)}) = 0$ for $j \neq i^*$. This simplifies the M-step: the mixture of GP posteriors~\eqref{eq:exma_m_step} reduces to a single GP posterior conditioned on the most likely candidate, and eliminates the need to track and update weights for all candidates. As a consequence, the computational complexity of the model estimation step is reduced from $O(N \cdot T^3)$ to $O(T^3)$ per iteration. While hard EM is computationally more efficient and requires less memory (no mixture representation), it is more prone to getting stuck in local optima. This typically happens because it commits to a single candidate at each iteration rather than maintaining uncertainty over multiple plausible candidates. The full EM approach is generally preferred when computational resources permit, as it better explores the posterior landscape. In our experiments, we employ the full EM algorithm as the computational cost remains manageable due to the moderate number of candidates used.

\subsection{Implementation Details}
\label{subsec:EM_implementation}
The EM algorithm detailed above can be integrated into an online calibration procedure. At each time step $T$, we apply a known input $u(T-1)$ and acquire a new Ronchigram observation $y(T)$. We preprocess the observation to obtain the latent representation $z(T) = \mu_\alpha(\mathcal{P}\{y(T)\})$. We then augment the dataset with the new latent space observation and input, updating the cumulative input $s(T) = \sum_{t=0}^{T-1} u(t)$. With the updated dataset $\{z(t), s(t)\}_{t=0}^{T}$, we run the EM iterations to refine the estimates of $x_0$ and $f$.

There are several reasons to acquire observations one at a time and to re-run EM with every new observation. First, it allows us to adaptively select the input $u(T)$ based on current estimates, improving convergence speed. Second, it allows us to select new candidate initial states based on the current posterior, reducing the total number of candidates while maintaining accuracy. Finally, it allows us to monitor convergence and stop the calibration process once satisfactory estimates are obtained, reducing unnecessary data collection.

\subsubsection{Input selection strategy}
The input selection strategy can significantly impact convergence speed and calibration accuracy. While random inputs can provide sufficient excitation for estimation, more sophisticated strategies can select inputs that maximize information gain about the unknown parameters. A simple acquisition heuristic is to select inputs that yield measurements at aberration locations where the GP posterior variance is highest, thereby reducing uncertainty in the latent mapping $f$. For computational efficiency, we employ a \emph{hard EM}-inspired approximation: rather than computing the full mixture posterior variance, we evaluate the variance using only the most likely candidate $i^* = \arg\max_i w_i$. The posterior variance for this candidate is
\begin{equation}
\label{eq:posterior_variance}
\text{Var}_{f|Z}^{(i^*)}(x) = k(x, x) - k_x^{(i^*)\top} (K^{(i^*)} + \Sigma_\nu I_{T+1})^{-1} k_x^{(i^*)},
\end{equation}
where $k_x^{(i^*)}$ is the cross-covariance vector from~\eqref{eq:single_gp_posterior}. This approximation reduces the computational cost from $O(N \cdot T^3)$ to $O(T^3)$ per variance query, making variance-based input selection practical. The approximation is justified when one candidate dominates the posterior (large $\max_i w_i$), as typically occurs after a few EM iterations. Inputs are then selected through
\begin{equation}
\label{eq:input_selection}
u(T) = \arg\max_{u \in \mathcal{U}} \text{Var}_{f|Z}^{(i^*)}(x_0^{(i^*)} + s(T) + u),
\end{equation}
where $\mathcal{U}$ is the set of allowable inputs (e.g., bounded defocus and astigmatism values). In practice, we optimize~\eqref{eq:input_selection} using grid search over a randomly sampled set of candidate inputs from $\mathcal{U}$.

\subsubsection{Candidate refinement}
In between acquiring new observations, we can refine the candidate set $\{x_0^{(i)}\}_{i=1}^{N}$ based on previous EM estimates to focus computational resources on high-probability regions while maintaining exploration. This refinement strategy is conceptually similar to resampling in particle filtering~\citep{Arulampalam2002}, where particles are redistributed based on their importance weights to concentrate samples in high-probability regions while avoiding sample degeneracy.

Our refinement strategy balances three objectives: exploiting the current posterior by retaining high-weight candidates, refining the search locally around promising regions, and maintaining global exploration to avoid premature convergence. Let $w_i$ denote the posterior weights from the most recent E-step. We construct the new candidate set by combining three components. First, we keep a fraction of the top-weighted candidates to preserve the current best estimates. Second, we generate new candidates by perturbing high-weight candidates with Gaussian noise, sampling parent candidates proportionally to their weights:
\begin{equation}
\label{eq:candidate_refinement}
x_0^{\text{new}} = x_0^{\text{parent}} + \mathcal{N}(0, \sigma^2 I_n),
\end{equation}
where the parent is drawn with probability proportional to $w_i$ and $\sigma > 0$ controls the exploration-exploitation tradeoff. Finally, we sample a small fraction of candidates uniformly over the aberration space to maintain coverage and prevent convergence to local optima. After refinement, the candidate weights are reset uniformly and the GP posteriors are recomputed in the next M-step.

\subsection{Algorithm summary}
The complete calibration procedure is summarized in Algorithm~\ref{alg:vae_em_calibration}. The VAE is trained offline on a dataset of simulated Ronchigram images. During online STEM calibration, the algorithm iteratively collects new Ronchigram observations, encodes them, and refines both the aberration estimate $x_0$ and the latent mapping $f$ using EM iterations.
\begin{algorithm}[t]
\caption{Iterative VAE-EM Calibration Algorithm}
\label{alg:vae_em_calibration}
\begin{algorithmic}[1]
\STATE \textbf{Input:} Trained VAE encoder $\mu_\alpha$, candidate set $\{x_0^{(i)}\}_{i=1}^{N}$, GP prior mean $\mu_f$ and kernel $k(\cdot, \cdot)$
\STATE \textbf{Output:} Estimated initial state $\hat{x}_0$, latent mapping $\hat{f}$
\STATE
\STATE Initialize $T = 0$, $s(0) = 0$
\STATE Acquire initial observation $y(0)$, compute $\tilde{y}(0) = \mathcal{P}\{y(0)\}$, encode $z(0) = \mu_\alpha(\tilde{y}(0))$
\REPEAT
    \STATE \textit{// EM iterations on dataset $\{z(t), s(t)\}_{t=0}^{T}$}
    \REPEAT
        \STATE E-step: Compute weights $w_i$ via~\eqref{eq:log_likelihood_gp_explicit} with $Z = [z(0), \ldots, z(T)]^\top$
        \STATE M-step: Update $\mu_{f|Z}(x)$ through~\eqref{eq:exma_m_step}
    \UNTIL{EM convergence}
    \STATE update candidate set $\{x_0^{(i)}\}_{i=1}^{N}$ using~\eqref{eq:candidate_refinement}
    \STATE Select next input $u(T)$ using~\eqref{eq:input_selection}
    \STATE $s(T+1) \gets s(T) + u(T)$, $T \gets T + 1$
    \STATE Acquire $y(T)$, compute $\tilde{y}(T) = \mathcal{P}\{y(T)\}$, encode $z(T) = \mu_\alpha(\tilde{y}(T))$
\UNTIL{convergence criterion met}
\end{algorithmic}
\end{algorithm}

\section{Results}
\label{sec:results}
We validate the proposed VAE-EM calibration method by applying it to real STEM data acquired on different days to capture day-to-day variations in system behavior. We compare the proposed method with existing automated calibration approaches in terms of calibration accuracy and convergence speed. In our previous work, we presented a detailed comparison of different Ronchigram-based calibration methods from the literature in terms of accuracy and speed. We refer the reader to that work for a comprehensive overview of existing approaches. Here, we focus on comparing our proposed VAE-EM method against the strongest baselines identified previously.

Our experiments consider three aberration parameters ($n=3$): defocus ($C1$) and two-fold astigmatism in both x- and y-directions ($A1x$, $A1y$). We use latent dimension $\ell = n = 3$ for the EM estimation experiments. For visualizations of latent spaces, we show results with $\ell = n = 2$ (defocus and x-astigmatism only).

\subsection{VAE captures aberration structure in latent space}
\begin{figure*}
    \centering
    \begin{minipage}{0.31\textwidth}
    \centering
    \includegraphics[width=\textwidth]{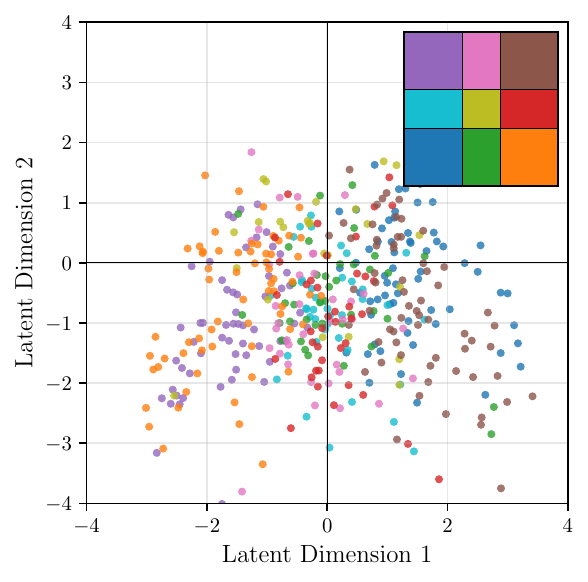}
    \end{minipage}
    \begin{minipage}{0.31\textwidth}
    \centering
    \includegraphics[width=\textwidth]{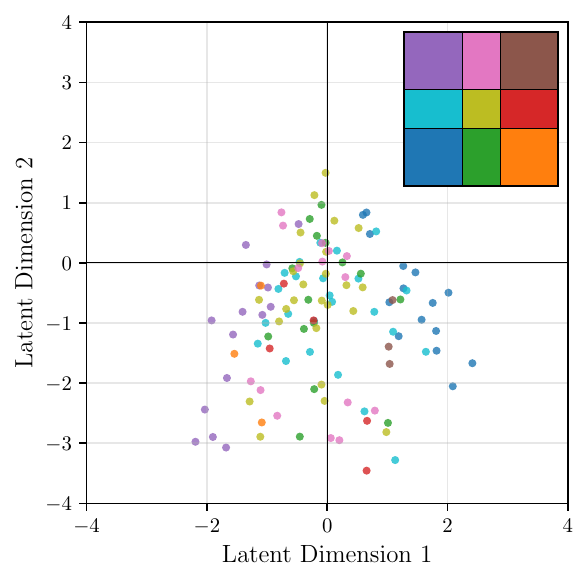}
    \end{minipage}
    \begin{minipage}{0.31\textwidth}
    \centering
    \includegraphics[width=\textwidth]{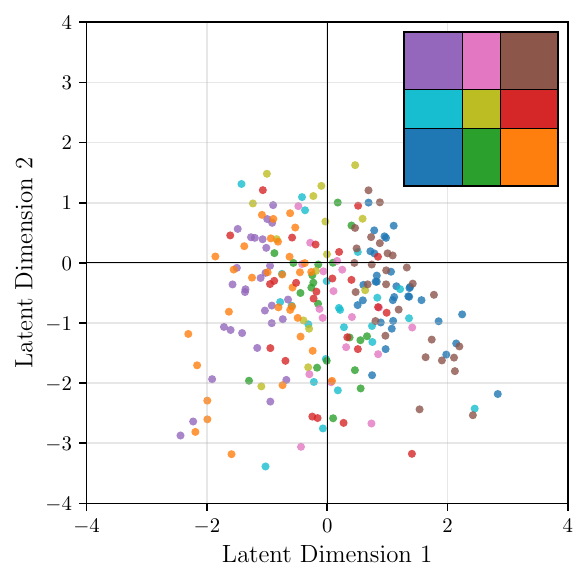}
    \end{minipage}
    \caption{Latent space distribution for Ronchigram datasets gathered on real STEM on different days. Left: dataset 1, middle: dataset 2, right: dataset 3. The latent space data points are colored according to their aberration values, with the color-coding scheme taken from Fig.~\ref{fig:images_grid}. We can observe consistency across datasets, as well as with respect to the simulated data from Fig.~\ref{fig:vae_latent_space_sim}, indicating that the VAE has learned a meaningful latent representation that reflects the underlying aberration states.}
    \label{fig:latent_space_distribution_experimental}
\end{figure*}

Fig.~\ref{fig:latent_space_distribution_experimental} shows the latent space distribution of the real STEM Ronchigram datasets using a 2D latent space for visualization.

We see consistency across the different datasets, as well as with respect to the simulated data from Fig.~\ref{fig:vae_latent_space_sim}. Observe that the experimental results are somewhat noisier and occupy a smaller region of the latent space compared to the simulated data. This is likely due to measurement noise and day-to-day variations in the microscope that are not captured by the simulator. Nevertheless, the overall structure is preserved: aberration states $x$ and $-x$ continue to map to similar latent locations (as seen by the color coding), while separation between different aberration quadrants is maintained. This indicates that the VAE has learned a meaningful representation that reflects the underlying physics.
\begin{figure*}
    \centering
    \includegraphics[width=0.8\textwidth]{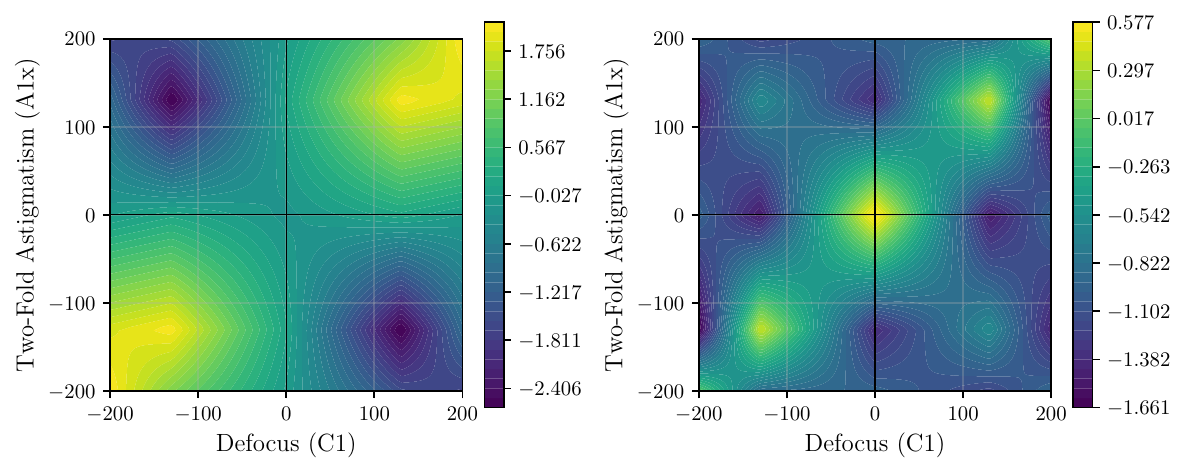}
    \caption{Fitted prior mean function $\mu_f(x)$ mapping aberration parameters to VAE latent space. Left: first latent dimension $z_1$ as a function of defocus ($C1$) and two-fold astigmatism ($A1x$). Right: second latent dimension $z_2$. The prior mean is generated by fitting piecewise linear bases to a large set of simulated Ronchigrams. Note the symmetry with respect to the origin, reflecting the physical property $f(-x) = f(x)$.}
    \label{fig:latent_maps}
\end{figure*}

From the latent space representation, we can fit the GP prior mean function $\mu_f(x)$ that maps aberration states to latent space dimensions. We use a grid of symmetrical piecewise linear bases, though other basis functions (e.g., radial basis functions, polynomials) could also be used. Fig.~\ref{fig:latent_maps} shows the fitted prior mean for the 2D latent space case. We see smooth mappings that capture the nonlinear relationship between aberration states and latent dimensions. These fitted prior means typically improve estimation accuracy for low numbers of observations, though we see limited improvements once enough data has been acquired.

\subsection{Superior calibration performance on real STEM data}
\begin{figure}
    \centering
    \includegraphics[width=0.5\textwidth]{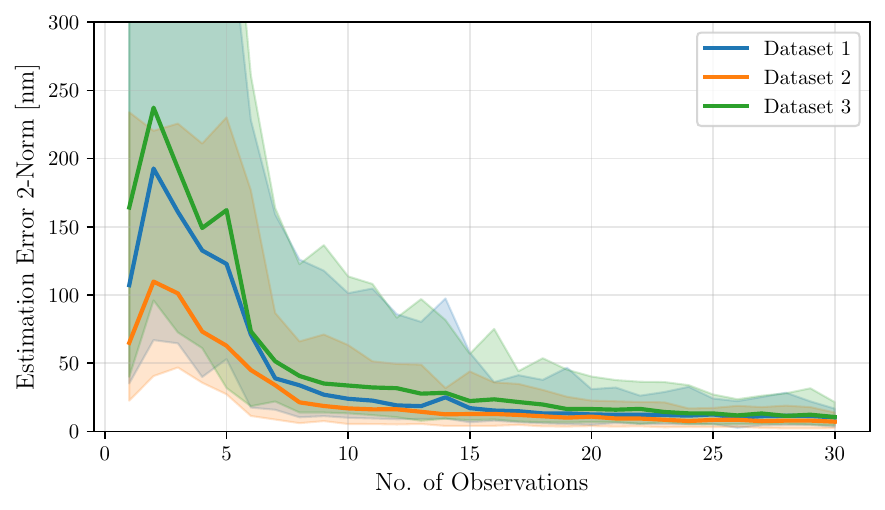}
    \caption{Estimation performance on real STEM data over 100 Monte Carlo runs with different random initializations with 3 aberration parameters ($C1, A1x, A1y$). We view the aberration estimation error 2-norm as a function of the number of Ronchigram observations used for calibration. The proposed VAE-EM method consistently outperforms existing automated calibration approaches in terms of speed and accuracy.}
    \label{fig:dataset_comparison}
\end{figure}

\begin{figure*}
    \centering
    \begin{minipage}{0.48\textwidth}
    \centering
    \includegraphics[width=\textwidth]{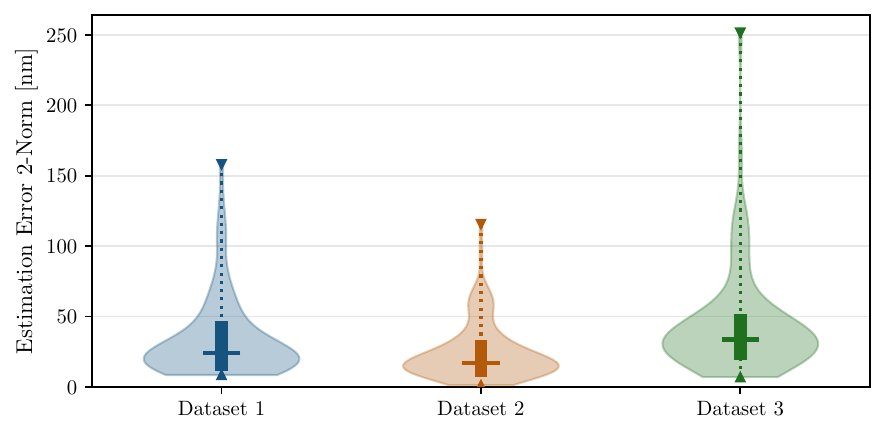}
    \end{minipage}
    \begin{minipage}{0.48\textwidth}
    \centering
    \includegraphics[width=\textwidth]{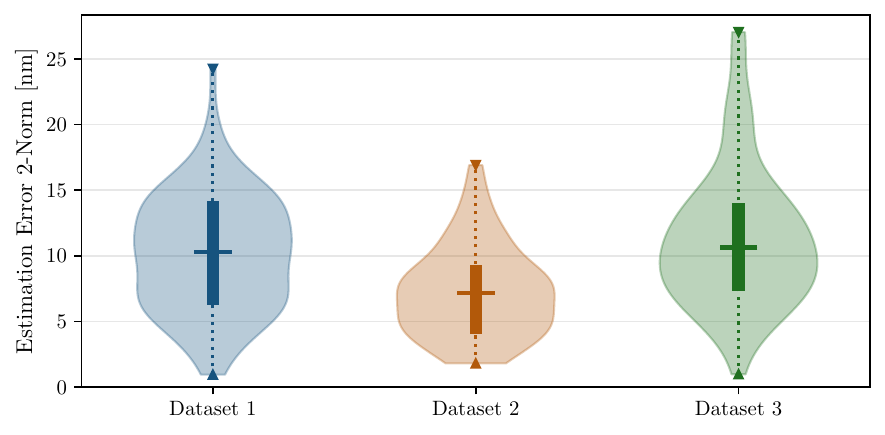}
    \end{minipage}
    \caption{Estimation error performance on real STEM data after 10 Ronchigram observations (left) and 30 observations (right). The violin plots summarize the distribution of estimation error 2-norms with 3 aberration parameters ($C1, A1x, A1y$) across 100 Monte Carlo runs with different random initializations.}
    \label{fig:zoomed_error_distribution}
\end{figure*}

Fig.~\ref{fig:dataset_comparison} and Fig.~\ref{fig:zoomed_error_distribution} show the aberration estimation error 2-norm as a function of the number of Ronchigram observations used for calibration, averaged over 100 Monte Carlo runs with different random initializations. After approximately 10 observations, the proposed VAE-EM method achieves the accuracy requirements for high-quality imaging (defocus and astigmatism below $10$ nm each) on over 50\% of runs. On all datasets, the proposed method converges after around 20 observations to very accurate estimates with high consistency across random initializations. The resulting calibration error 2-norms are reduced by over a factor of 2 compared to our previous Bayesian optimization approach~\citep{VanHulst2025} on the same datasets. Our previous work established that this Bayesian optimization approach already matched the performance of existing automated methods on real STEM data. Current commercial solutions such as CEOS and OptiSTEM(+) typically require approximately one minute to reach acceptable calibration specifications. In contrast, our VAE-EM method requires only around 20 Ronchigrams, corresponding to roughly 20 seconds of acquisition time, while achieving substantially higher accuracy.

\section{Discussion}
\label{sec:discussion}
We have presented a VAE-EM approach for automated STEM calibration that addresses two fundamental limitations of existing data-driven methods: the reliance on scalar image representations and the simulation-to-reality gap. Multi-dimensional VAE representations preserve richer information about the aberration state than scalar quality metrics, which rapidly lose information content as the number of calibration parameters increases. By jointly estimating both the calibration state and the latent mapping using EM, we achieve over $2\times$ more accurate calibration compared to state-of-the-art methods while requiring fewer observations.

The global identifiability result (Theorem~\ref{thm:identifiability}) provides theoretical grounding for our approach by ensuring that the joint estimation problem has a unique solution under mild conditions on the basis functions. This is notable because standard EM approaches for inverse problems often lack such guarantees~\citep{Wu1983,Tarantola2005}. Our approach provides an elegant way of selecting these basis functions: they are generated by a symmetrized kernel, which in the limit can approximate any even function arbitrarily well. Prior knowledge can be embedded through the choice of base kernel, prior mean, and hyperparameters.

\subsection{Limitations and Future Work}
A limitation of our approach is the need for substantial training data to train the VAE. We resolve this by using a digital twin (simulator) and explicitly learning the resulting model mismatch through the GP posterior. However, the simulated dataset must cover essentially all aberration combinations that we aim to estimate in order to train a sufficiently expressive VAE. For practical deployment, we recommend using iterative candidate refinement and variance-based input selection. These strategies significantly reduce computational cost while maintaining calibration accuracy. If further reduction is needed, hard EM can be employed at the cost of some estimation accuracy.

Generalization to higher-order aberrations is conceptually straightforward since the evenness constraint still holds, though practical challenges arise from GP computation scaling and the curse of dimensionality. A hierarchical approach that first calibrates lower-order aberrations before refining higher-order terms may address these challenges. Additionally, higher-order aberrations tend to have localized effects on Ronchigrams, which may require patching the Ronchigrams and applying Fourier transforms locally to extract relevant latent features.

\subsection{Applicability Beyond STEM}
While developed for STEM calibration, the methods presented here may extend to other inverse problems. We now outline when the framework applies.

The starting point is an inverse problem where high-dimensional observations arise from low-dimensional parameters. In our case, Ronchigrams contain approximately $10^5$ pixels yet are determined by a few aberration parameters. This gap enables a VAE to learn a compressed representation without losing estimation-relevant information. When training data comes from a simulator rather than the real system, the learned latent mapping will not perfectly match reality. This simulation-to-reality gap means the VAE representations are biased. Joint estimation of both the state and the residual model mismatch using EM can correct this bias, but such joint estimation is generally ill-posed: many combinations of state and model parameters may explain the observations equally well.

Structural constraints can restore identifiability. In our case, the evenness constraint from Fourier optics allows us to construct a symmetry-constrained GP kernel, which establishes global identifiability (Theorem~\ref{thm:identifiability}). Non-injectivity of the observation mapping is not itself a requirement. Rather, the particular structure of our non-injectivity enables us to recover identifiability. An injective observation mapping would trivially ensure identifiability and also fit within this framework. The framework also requires that the state dynamics equations are known and invertible. Our state-space model has trivial dynamics where the next state equals the current state plus a known input, allowing all observations to be related back to a common initial state for joint estimation.

When the VAE is trained on representative real data, the simulation-to-reality gap becomes negligible. The proposed framework then reduces to pure state estimation in the learned latent space, with the latent mapping $f$ estimated directly from training data rather than learned online. In essence, our framework encompasses both joint state-model estimation and pure VAE-based state estimation as a special case.

Several application domains may fit the VAE-EM framework. Optical systems with known symmetries, such as adaptive optics for telescopes or microscopy, share the evenness structure that enables our identifiability result. Phase retrieval problems exhibit similar non-injectivity due to the loss of phase information. Medical imaging modalities where simulators exist but patient-specific variations cause model mismatch, such as MRI or CT reconstruction, may also benefit from joint state-model estimation. For applications where representative real data is available, such as industrial quality control with abundant labeled samples, the simpler VAE-based state estimation applies directly without requiring the EM iterations for model learning.

\subsection{Conclusion}
\label{sec:conclusion}
This work introduced a VAE-EM framework that surpasses state-of-the-art STEM calibration performance by combining multi-dimensional image representations with joint state-model estimation. The key theoretical insight is that structural constraints, such as the evenness property from Fourier optics, can restore global identifiability in joint estimation problems that are otherwise ill-posed. This provides a principled approach to bridging simulation-to-reality gaps in data-driven inverse problems.

\section*{Acknowledgments}
The authors would like to acknowledge the work of Joyce de Heer and Tim Jansen, who have contributed to this research as part of their graduation projects at Thermo Fisher Scientific.

The authors would also like to thank Thermo Fisher Scientific for providing access to the real STEM set-up and STEM simulator. In particular, we would like to thank Maurits Diephuis, Narges Javaheri and Maurice Peemen for their support.

\section*{Declaration of generative AI and AI-assisted technologies in the manuscript preparation process}
During manuscript preparation, the authors used Claude Sonnet/Opus (Anthropic) to aid development of the analysis code and to perform editing and grammar enhancements in the manuscript. After using this tool, the authors reviewed and edited the content as needed and take full responsibility for the content of the manuscript.

\bibliographystyle{IEEEtran}
\bibliography{references}

\begin{IEEEbiography}[{\includegraphics[width=1in,height=1.25in,clip,keepaspectratio]{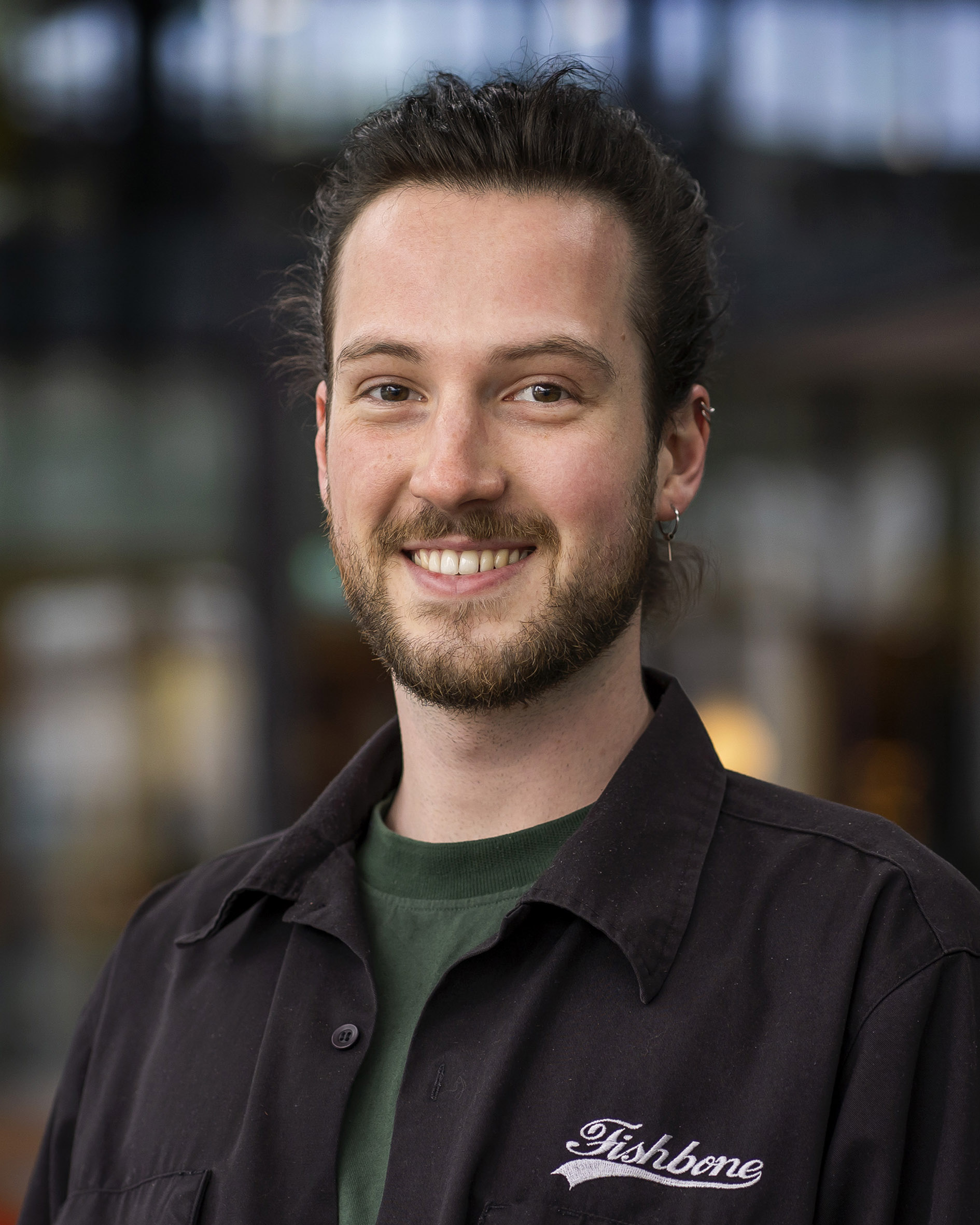}}]{Jilles S. van Hulst} received the M.Sc. degree in mechanical engineering from the Eindhoven University of Technology (TU/e) in 2022, where he is currently pursuing a Ph.D. degree under the supervision of Prof. W.P.M.H (Maurice) Heemels and Prof. Duarte J. Antunes within the Control Systems Technology Section. His research interests include Bayesian statistics and data-driven estimation and control.
\end{IEEEbiography}

\begin{IEEEbiography}[{\includegraphics[width=1in,height=1.25in,clip,keepaspectratio]{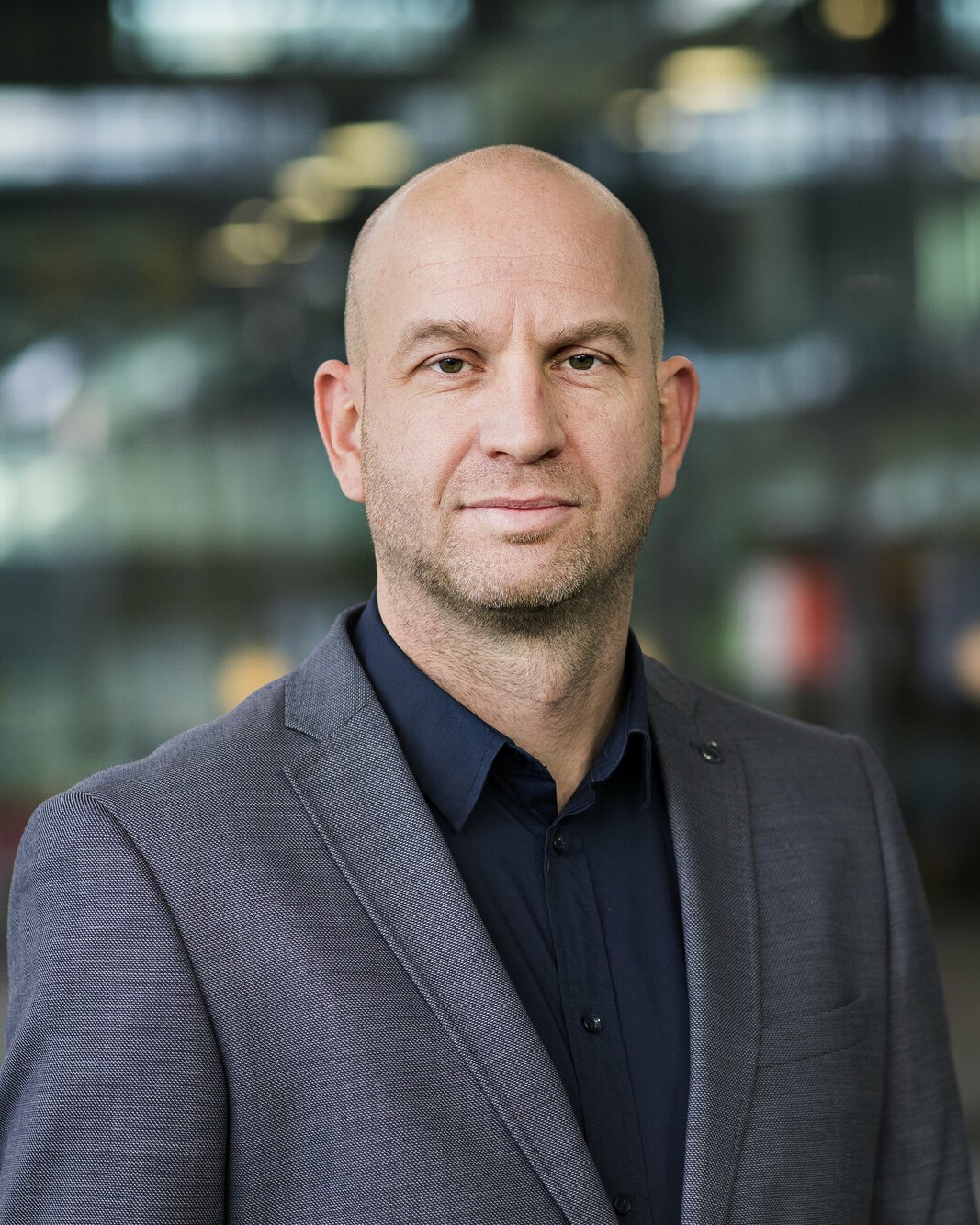}}]{W.P.M.H. (Maurice) Heemels} received M.Sc. (Mathematics) and Ph.D. (EE, control theory) degrees (summa cum laude) from the Eindhoven University of Technology (TU/e) in 1995 and 1999, respectively. From 2000 to 2004, he was with the Electrical Engineering Department, TU/e, as an assistant professor, and from 2004 to 2006 with the Embedded Systems Institute (ESI) as a Research Fellow. Since 2006, he has been with the Department of Mechanical Engineering, TU/e, where he is currently a Full Professor and Vice-Dean. He held visiting professor positions at ETH, Switzerland (2001), UCSB, USA (2008) and University of Lorraine, France (2020). He is a Fellow of the IEEE and IFAC, and the chair of the IFAC Technical Committee on Networked Systems (2017–2023). He served/s on the editorial boards of Automatica, Nonlinear Analysis: Hybrid Systems (NAHS), Annual Reviews in Control, and IEEE Transactions on Automatic Control, and is the Editor-in-Chief of NAHS as of 2023. He was a recipient of a personal VICI grant awarded by NWO (Dutch Research Council) and recently obtained an ERC Advanced Grant. He was the recipient of the 2019 IEEE L-CSS Outstanding Paper Award and the Automatica Paper Prize 2020–2022. He was elected for the IEEE-CSS Board of Governors (2021–2023). His current research includes hybrid and cyber–physical systems, networked and event-triggered control systems and model predictive control.
\end{IEEEbiography}

\begin{IEEEbiography}[{\includegraphics[width=1in,height=1.25in,clip,keepaspectratio]{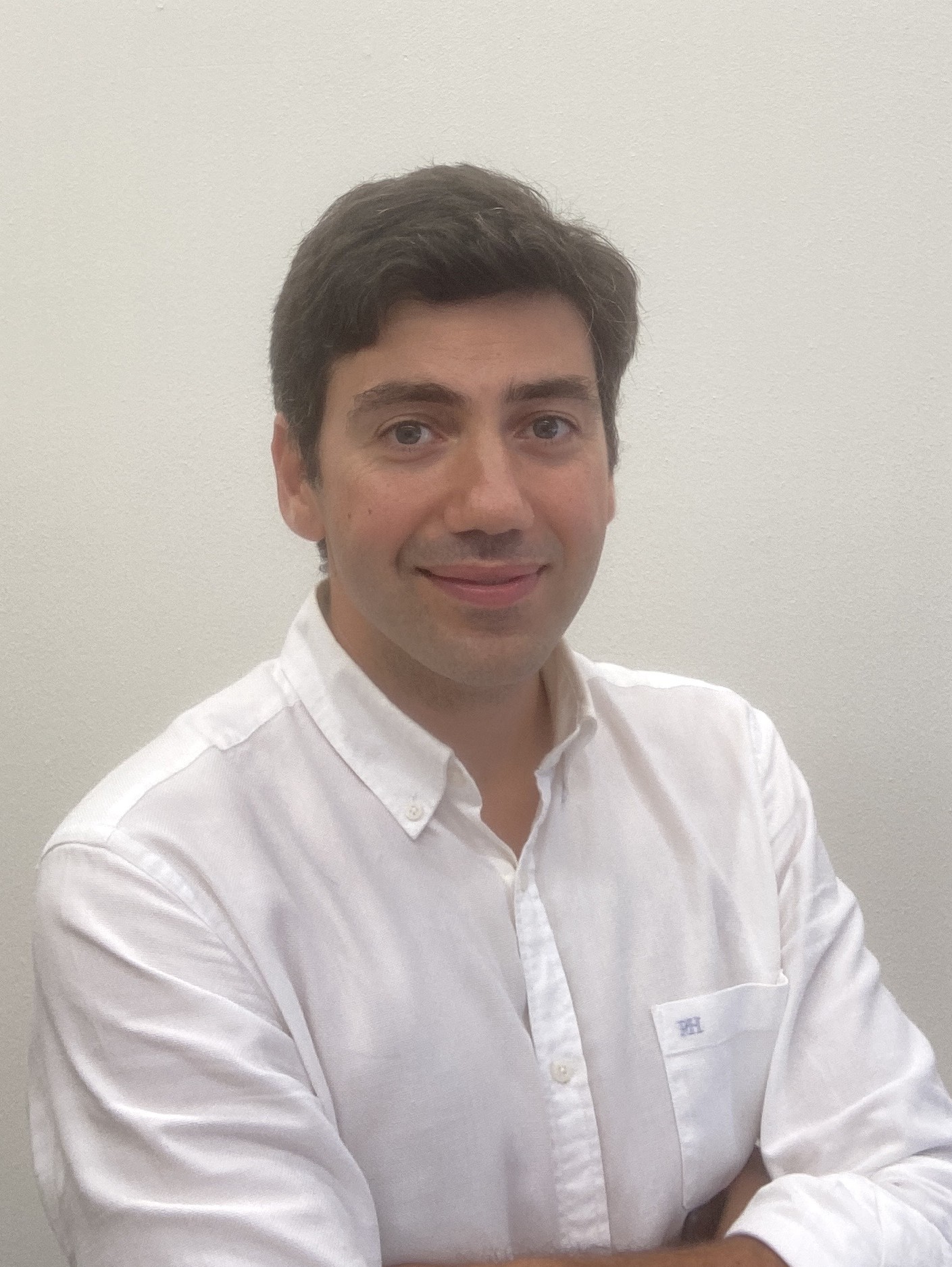}}]{Duarte J. Antunes} obtained the Ph.D. degree (cum laude) in automatic control from the Institute for Systems and Robotics, IST, Lisbon, in 2011, in collaboration at the University of California, Santa Barbara, CA, USA. From 2011 to 2013, he held a postdoctoral position with the Eindhoven University of Technology (TU/e) and became an Assistant Professor in 2013. He is currently an Associate Professor with the Department of Mechanical Engineering, TU/e. His research interests include networked control systems, stochastic control, approximate dynamic programming, and robotics.
\end{IEEEbiography}

\end{document}